\documentclass{article} 
\usepackage{iclr2025_conference,times}

\usepackage{hyperref}
\usepackage{url}

\usepackage{xcolor,soul,framed} 
\usepackage{multirow}
\colorlet{shadecolor}{yellow}
\usepackage[pdftex]{graphicx}
\DeclareGraphicsExtensions{.pdf,.jpeg,.png}

\usepackage[cmex10]{amsmath}

\usepackage{array}
\usepackage{mdwmath}
\usepackage{mdwtab}
\usepackage{eqparbox}
\usepackage{url}
\usepackage{booktabs}
\usepackage{mathstuff}
\usepackage{tikz}
\usetikzlibrary{positioning,fit,shapes}

\usepackage{amsthm}
\usepackage{hyperref}
\usepackage{accents}
\usepackage{subcaption}

{
  \theoremstyle{plain}
  
}

\def\ki{Department of Physiology and Pharmacology, Karolinska Institutet, Stockholm, Sweden}
\def\kth{Division of Information Science and Engineering, KTH, Stockholm, Sweden}
\def\waspsentence{This work was partially supported by the Wallenberg AI, Autonomous Systems and Software Program (WASP) funded by the Knut and Alice Wallenberg Foundation. The computations were enabled by the Berzelius resource provided by the Knut and Alice Wallenberg Foundation at the National Supercomputer Centre.}

\title{A Matrix Variational Auto-Encoder for\\Variant Effect Prediction in Pharmacogenes}

\author{Antoine~Honoré$^{1}$\thanks{Corresponding author: \href{mailto:antoinehonor@gmail.com}{antoinehonor@gmail.com}. \waspsentence} , Borja Rodr\'iguez G\'alvez$^{3}$, Yoomi Park$^{2}$, Yitian Zhou$^{2}$,\\ \textbf{Volker M. Lauschke}$^{2}$\textbf{, Ming Xiao}$^{1}$\\
$^{1}$ \kth\\
$^{2}$ \ki\\
$^{3}$ Independent researcher
}

\def\tablescale{0.8}

\iclrfinalcopy 
\begin{document}

\maketitle

\begin{abstract}
Variant effect predictors (VEPs) aim to assess the functional impact of protein variants, traditionally relying on multiple sequence alignments (MSAs). This approach assumes that naturally occurring variants are fit, an assumption challenged by pharmacogenomics, where some pharmacogenes experience low evolutionary pressure. Deep mutational scanning (DMS) datasets provide an alternative by offering quantitative fitness scores for variants. In this work, we propose a transformer-based matrix variational auto-encoder (matVAE) with a structured prior and evaluate its performance on 33 DMS datasets corresponding to 26 drug target and ADME proteins from the ProteinGym benchmark. Our model trained on MSAs (matVAE-MSA) outperforms the state-of-the-art DeepSequence model in zero-shot prediction on DMS datasets, despite using an order of magnitude fewer parameters and requiring less computation at inference time. We also compare matVAE-MSA to matENC-DMS, a model of similar capacity trained on DMS data, and find that the latter performs better on supervised prediction tasks. Additionally, incorporating AlphaFold-generated structures into our transformer model further improves performance, achieving results comparable to DeepSequence trained on MSAs and finetuned on DMS. These findings highlight the potential of DMS datasets to replace MSAs without significant loss in predictive performance, motivating further development of DMS datasets and exploration of their relationships to enhance variant effect prediction.
\end{abstract}

\section{Introduction}
Variant effect predictors (VEPs) are mathematical models aiming at predicting the effect of one or multiple variants in a sequence of amino-acids (AAs).
The effect of a protein variant is typically defined as a loss or gain of function of a cell carrying the variant, compared to a cell carrying a wild-type (WT) protein without variant.
The accurate prediction of variant effect has many promising applications for personalized medicine, particularly in the field of pharmacogenomics, where variants on drug targets or Absorption-Distribution-Metabolism-Excretion (ADME) proteins are of particular interest \citep{huangComingAgeNovo2016}. 
In this context, VEPs can be used to assess individual patient response to chemotherapeutic treatments from their genetic background, thus eliminating the need for multiple attempts at treatments.
The most effective VEPs have been designed using data from multiple sequence alignments (MSAs) and based on the \textit{conservation assumption}: fit variants were selected out by nature and thus, learning a distribution over variants found in nature implicitly captures the biochemical constraints that characterize fit variants.
New sequencing techniques combined with machine learning could lead to significant advances in variant effect prediction, by providing quantitative data in region of the protein sequence space unexplored in existing MSA datasets.
Deep mutational scanning (DMS) has recently emerged as a way to yield large-scale datasets of protein quantitative fitness scores \citep{fowlerDeepMutationalScanning2014}. 
The fitness scores can also be obtained with different selection assays, allowing to quantify various effects, e.g. effect on phenotype or effect on structure.
DMS thus allows to challenge the conservation assumption of VEPs design from MSAs.
This is in turn is of particular importance in pharmacogenomics, since pharmacogenes are generally under low evolutionary pressure \citep{zhouChallengesOpportunitiesAssociated2022,ingelman-sundbergIntegratingRareGenetic2018}.

In this article, we design a VEP and use it to evaluate the validity of the conservation assumption for pharmacogene-related proteins.
Our architecture exploits the structure of variational auto-encoders (VAEs) and allows models of similar capacity and designs to be trained on both MSA and DMS data.
We also exploit a transformer architecture in order to improve upon existing VAE-based VEPs.
Both VAE and transformers are key components of the best performing models in the ProteinGym benchmark \citep{notinProteinGymLargeScaleBenchmarks2023}.
We experiment with a VAE-based model exploiting multimodal priors, and we derive a matrix encoding scheme inspired from linear matrix decomposition to replace the input flattening operation found for instance in DeepSequence.

\paragraph{Contributions}
Our contributions are summarized as follows: \begin{enumerate}
    \item We design protein specific models combining a VAE and a transformer for variant effect prediction. We study their zero-shot prediction performances on $33$ deep mutational scanning (DMS) datasets of drug related and ADME proteins available in the ProteinGym benchmark.
    \item We propose a new way to train VAEs for deep unsuperivsed clustering using expressive latent prior based on discrete distributions. We evaluate the model on zero-shot predictions when trained on MSA data available in ProteinGym. We experiment with and without adding protein structure information to our model.
    \item We adapt our model to directly predict labels from DMS datasets using a prediction head from the latent space, thus preserving our model capacity. In light of the comparison in performances of the models trained unsupervised on MSA and supervised DMS label data, we discuss the extent of the validity of the conservation assumption.
\end{enumerate}

\subsection{Related works}
\paragraph{Zero-shot predictors}
VEPs exploiting site-independent position-wise frequencies of AAs in MSAs remain the methods of choice in pharmacology, e.g. SIFT or Polyphen-2 \citep{ngSIFTPredictingAmino2003,adzhubeiMethodServerPredicting2010b,durbin1998biological}.
However, other models can achieve much better zero-shot prediction performances on at least one pharmacogene-related protein DMS dataset (Details in Table~\ref{tab:best_model_family}), according to the recent ProteinGym variant effect prediction benchmark \citep{notinProteinGymLargeScaleBenchmarks2023}.
Many of these models compute the functional cellular effect of a variant $v$ compared to a wild-type sequence ${\umbx}^{(wt)}$, via the log-likelihood ratio:\begin{equation}\label{eq:logratio}
    \hat{y}=\ln \frac{p({\umbx}^{(v)})}{p({\umbx}^{(wt)})},
\end{equation}
where $p(.)$ is a generative probability density chosen to maximize $p({\umbx}^{(v)})$, for sequences $\umbx^{(v)}$ from the MSA.
For instance, the Evolutionary Scale modeling (ESM) approaches \citep{rivesBiologicalStructureFunction2021}, rely solely on a transformer-based protein language model (PLM) for modeling the distribution over sequences in MSAs.
Tranception \citep{notinTranceptionProteinFitness2022} additionally integrates predictions using position-wise frequencies of AAs in MSAs.
TranceptEVE \citep{notinTranceptEVECombiningFamilyspecific2022} combines the Tranception model with a VAE-based model \citep{frazerDiseaseVariantPrediction2021} for AA sequence modeling.
Other methods such as Masked Inverse Folding (MIF) \citep{yangMaskedInverseFolding2023} learn to predict protein sequences from a given structure. 
VESPA \citep{marquetEmbeddingsProteinLanguage2022} combines protein sequence embedding from PLMs with known bio-mechanical properties of AAs to predict variant effect with a linear regression model.
Other model do not rely on the ratio in \eqref{eq:logratio} to compute variant effect.
MSA Transformer \citep{raoMSATransformer2021a} is based on ESM and uses axial attention to optimize a masking loss over an entire MSA, rather than on individual sequences.
It learns a representation of Hamming distances in the MSA and the hamming distance to WT sequence is used as a proxy for variant effect.
GEMME \citep{laineGEMMESimpleFast2019} predicts variant effect via the distance to WT sequence in an evolutionary tree.
This approach shows very good performances and has several order of magnitude fewer parameters than transformer-based approaches.
DeepSequence \citep{riesselmanDeepGenerativeModels2018} introduces a VAE and approximates the distribution of input data $\umbx$ (Eq. \eqref{eq:logratio}) with the variational evidence lower bound.

\paragraph{Supervised learning predictors}
Recently, several models combining DMS and MSA datasets have been proposed \cite{hsuLearningProteinFitness2022a}.
The general idea is to combine sequence embeddings, e.g. sequence one-hot encoding, with evolutionary fitness scores from pretrained models such as ESM or DeepSequence.
ProteinNPT is a conditional pseudo-generative model designed for exploiting DMS data, jointly with MSA data in a semi-supervised framework \citep{notin2023proteinnpt}.
In addition to their novel architecture, the authors introduce several baselines consisting in exploiting prediction scores from zero-shot prediction models pretrained on MSA, including DeepSequence and MSA Transformer.
SPIRED is a recent framework able to predict fitness scores as well as protein structure \citep{chenEndendFrameworkPrediction2024}. 
A pretrained ESM model is used for sequence embedding, and graph attention networks and multilayer perceptron are trained using DMS data in a supervised framework.

\paragraph{Multi-modal prior distributions for VAEs}
VAEs, e.g. DeepSequence, assume that the input data $\umbx\in\{0,1\}^{L \times d}$ are generated from a latent variable of a $D$-dimensional vector space: $\mbz\in\R^D$.
The latent variable is assumed drawn from a Gaussian prior $p(\mbz)$, and the generative process is modeled with a distribution $p_\theta(\umbx|\mbz)$.
The explicit modeling of the latent variable $\mbz$ is an interesting feature of VAEs, because it allows to put a formal prior distribution on the latent space.
The other mentioned models do not impose such structure, although interestingly the learnt representations in ESM was shown to correlate with known bio-mechanical properties of AAs \citep{rivesBiologicalStructureFunction2021}.
Multimodal mixture of Gaussian (MOG) priors have been proposed as latent prior distributions for unsupervised clustering tasks \citep{dilokthanakulDeepUnsupervisedClustering2017} with VAE. 
The authors used trainable mean and covariances in latent space and showed through data sampling that the learnt mixture components corresponded to meaningful characteristics of the input data. 
This was shown to have potential implications for model interpretability in biological contexts \citep{varolgunesInterpretableEmbeddingsMolecular2020}.
Further, the VampPrior has been designed so that the statistics of the mixture components explicitly depend on input space prototypes \cite{tomczakVAEVampPrior2018}.
This provides meaningful variables to probe for interpretability rather than using a sampling scheme.
Bayesian VAEs such as DeepSequence were shown to have good performances as VEP. 
However, they require a lot of parameters (~30M) and a lot sampling in order to get good results. This makes them difficult to use as expert model in an ensemble for instance. 
To the best of the authors knowledge, current methods employing VAEs for VEP have only been designed with unimodal prior distributions.
We propose to structure the latent with an implicit mixture of discrete distributions, which is readily interpretable, reduces the number of parameters, and does not require sampling at inference.


\section{Methods}
A detailed description of the matVAE-MSA architecture is provided in section \ref{sec:modeldescription}. 
A reduction of the architecture with similar capacity and that can be trained on DMS data: matENC-DMS, is proposed in section \ref{sec:modelreduction}.
The datasets that are used to train and evaluate the models are introduced in section \ref{sec:dataset}. 

\begin{figure}[ht!]
    \begin{center}
    \begin{tikzpicture}[font=\footnotesize,>=stealth,auto,node distance=1.6]
    
    \node (input) [rectangle,minimum width=1.2cm] {Input};
    \node (trans_input) [rectangle, draw, right=of input] {Transformer};
    \node (mlpin) [rectangle, draw, right=of trans_input] {DwFC};
    \node (dummy) [rectangle, below=0.8cm of mlpin] {};
    \node (CE) [rectangle, draw, below=0.6cm of input] {$\text{CE}(\umbx,\hat{\umbx})$};
    \node (vae) [rectangle, draw, right=of dummy] {FCB};
    \node (mlpout) [rectangle, draw, below=0.8cm of dummy] {DwFC};
    \node (trans_output) [rectangle, draw, left=of mlpout] {Transformer};
    \node (output) [rectangle, left=of trans_output,minimum width=1.2cm] {Output};
    
    \draw[->,thick] (input) -- (trans_input) node[midway, above] {$\umbx \in \mathbb{R}^{L \times d}$};
    \draw[->,thick] (trans_input) -- (mlpin) node[midway, above] {$\umbx' \in \mathbb{R}^{L \times d}$};
    \draw[->,thick] (mlpin.east) -| (vae.north)  node[midway, above] {$\umbs \in \mathbb{R}^{H\times d}$};
    \draw[->,thick] (vae.south) |- (mlpout.east) node[midway, above left] {$\hat{\umbs} \in \mathbb{R}^{H\times d}$};
    \draw[->,thick] (mlpout) -- (trans_output) node[midway, above] {$\hat{\umbx}'\in\mathbb{R}^{L \times d}$};
    \draw[->,thick] (trans_output) -- (output) node[midway, above] {$\hat{\umbx}\in \mathbb{R}^{L \times d}$};
    \draw[->,thick] (output) -- (CE) node[midway, above] {};
    \draw[->,thick] (input) -- (CE) node[midway, above] {};
    \node[draw, dashed, fit=(trans_input) (mlpin),minimum height=1.4cm] (fit) {};
    \node[above=0cm of fit] (fit_text) {Matrix encoding};
    \end{tikzpicture}
    
    \caption{Model architecture of matVAE-MSA. DwFC: Dimensionwise fully connected layer. CE: Cross-entropy loss. FCB: Fully connected bottleneck.}
    \label{fig:overallmodel}
    \end{center}
\end{figure}
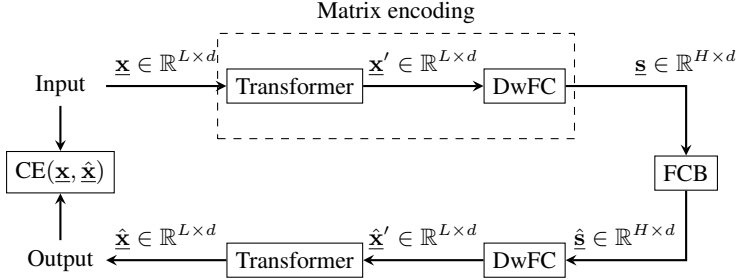
\subsection{Model description}\label{sec:modeldescription}

\subsubsection{Matrix decomposition and encoding}\label{sec:qr}
Matrix decomposition is a set of methods in linear algebra consisting in decomposing a matrix $\umbx\in\R^{L\times d}$ (here $d<L$), into two (or more) matrices with interesting structure, e.g. unitary, triangular, diagonal. 
For instance, the QR decomposition of $\umbx\in\R^{L\times d}$ is of the form
$\umbx=\mbQ[\mbR^T,\mathbf{0}^T]^T$, where $\mbQ\in\R^{L\times L}$ is unitary, $\mbR\in\R^{d\times d}$ is upper triangular and $\mathbf{0}$ is a $(L-d\times d)$-dimensional matrix of zeros.
We call matrix encoding the use of the low dimension factor, here the $\mbR$ matrix, as a compressed representation of the input $\umbx$.
For QR decomposition, the matrix encoding can be formulated: \begin{equation}\label{eq:matrixencoding}
    [\umbs^T,\mathbf{0}^T]^T = \mbW \umbx \in \R^{L\times d},
\end{equation}
where $\mbW = \mbQ^{-1} \in \R^{L\times L}$ is a linear transform and $\umbs=\mbR$.
Linear decomposition methods are often related to singular value decomposition, for instance the diagonal elements of $\mbR$ in the QR decomposition are the singular values of $\umbx$. 
However, for one-hot encoded sequences of AAs, the singular values are counts of each AA in the sequence, and the singular vectors are a permutation of $\mathbb{I}_d$ determined by the ordering of the counts of the AAs.
In other words, the encoding produced with such linear methods does not account for the global or relative position of AAs in the sequence, and in particular, randomly permuting the rows of $\umbx$ leads to the same encoding.
To ensure that the model is flexible enough to learn a useful encoding, we propose to learn a representation of $\umbx$ with a transformer, prior to reducing the first dimension to a fixed $H < L$ with a trainable linear transform. 
This is formulated as follows: \begin{align*}
    \umbx' &= \text{Transformer}( \umbx ) \in \R^{L\times d},\\
    \umbs &= \text{DwFC}( \umbx' ) \in \R^{H\times d},
\end{align*}
where $\text{Transformer}(.)$ and $\text{DwFC}(.)$ are specified in the paragraphs below. 

\paragraph{Transformer}\label{sec:transformer}
Transformers are effective sequence models that can transfer information between any two positions within a sequence.
The model we use is similar to the multi-layer encoding transformer in \citep{vaswaniAttentionAllYou2017}. 
Individual layers encode an input sequence $\umbx\in \R^{L\times d}$ into a sequence $\umbx'\in \R^{L\times d}$ as follows: \begin{equation}\label{eq:transformerlayer}\begin{aligned}
    \umbx_1 &= \text{Norm}(\tau_1\umbx + (1-\tau_1)\text{Attn}(\umbx)),\\
    \umbx' &= \text{Norm}(\tau'\umbx_1 + (1-\tau')\text{FC}(\umbx_1)),
\end{aligned}\end{equation}
where $\text{Norm}(.)$ is a Layer Normalization \citep{baLayerNormalization2016}, 
$\text{FC}(.)$ is a fully connected (FC) network with ReLU activations, 
and $\text{Attn}(.)$ is the masked scaled dot product attention from \citep{vaswaniAttentionAllYou2017}.
We use trainable $\tau_1,\tau'\in[0,1]^2$ to control the gradient flow in the gated skip connections \citep{heIdentityMappingsDeep2016}.
Note that we do not use positional encoding since \citep{rivesBiologicalStructureFunction2021} showed that PLMs did not necessarily benefit from it. Instead, structure information is encoded in the mask of the attention layer (See section \ref{sec:hyperparams}).

\paragraph{Dimension-wise FC (DwFC)}\label{sec:dwmlp}
We call DwFC the FC linear layer inspired from Eq. \ref{eq:matrixencoding}.
This layer is similar to a flattening followed by a linear transform with bias, but requires less parameters since the same linear transform is used across dimensions. 
This operation replaces the direct flattening of the input in DeepSequence.
$\umbx'\in \R^{L\times d}$ is encoded in a protein length independent representation $\umbs\in \R^{H\times d}$ as follows:
\begin{align}\label{eq:DwMLP}
    \umbs = \mbU \umbx' + \mbb,
\end{align}
where $\mbU\in\R^{H\times L}$ and $\mbb\in\R^{H}$ are trainable weight and bias parameters. 


\subsubsection{Fully connected bottleneck (FCB) with Dirichlet latent prior}\label{sec:fcb}
We extend the work carried out in DeepSequence, by using the simplex $S^{D}=\{\mbz\in[0,1]^D|\sum_i z_i=1\}$ as latent space. 
The FCB is similar to a classical VAE \citep{kingmaAutoEncodingVariationalBayes2014}. 
The input $\umbs\in\R^{H\times d}$ is flattened and then encoded into a latent representation of fixed dimension:
\begin{equation}\label{eq:hvariable}
    \mbh=\text{softmax}(\text{FC}(\text{Vec}(\umbs)))\in\R^D,
\end{equation}
where $\text{Vec}(.)$ denotes the flattening operation and $\text{softmax}(.)$ ensures that the vector $\mbh$ represents a discrete probability distribution.
For training, we introduce robustness by drawing the latent vector $\mbz\sim \text{GumbelSoftmax}(\mbh)$ using the reparameterization trick \citep{potapczynskiInvertibleGaussianReparameterization2020}.
At test time, we use $\mbz=\mbh$ to reduce stochasticity.
The latent vector $\mbz\in\R^{D}$ is then used as input to a decoder network that aims to reconstruct the input $\umbs$. 
The output to the decoding part of the FCB, is denoted $\hat{\umbs} \in\R^{H\times d}$.

\subsubsection{Decoding}
The decoding process denoted $p_\theta(\umbx|\mbz)$, is symmetrical to the encoder, i.e. consists of a decoding FC layer, and a dimension wise FC layer followed by a transformer.
One important difference is that the decoding transformer includes a temperature softmax output operation to ensure that the rows of the reconstructed $\hat{\umbx}\in\R^{L\times d}$ define proper discrete distributions.

\subsubsection{Loss function}
Our loss function is directly derived from the negative evidence lower bound (ELBO) used in VAEs \citep{kingmaAutoEncodingVariationalBayes2014}:
\begin{equation}\label{eq:elbo}
\text{ELBO}(\umbx;\phi,\theta) = D_{KL}(q_\phi(\mbz|\umbx) || p(\mbz)) + \mathbb{E}q_\phi[\log p_\theta(\umbx|\mbz)],
\end{equation}
where on the RHS, the first term is the KLD between the approximated posterior distribution $q_\phi$, and the prior $p$, and the other is an expected reconstruction loss (defined in section \ref{sec:modeltraining}).

\textbf{Prior}: Our prior distribution is a mixture of $D$ degenerate discrete distributions described by one-hot vectors. 
However, KL divergence with a mixture distributions does not exist in closed form. Therefore, formulating it explicitly would require to use a bound instead.
In addition it would require specifying or learning statistics of the mixture.
Instead, we replace the explicit KL divergence minimization in Eq.~\eqref{eq:elbo}, with the minimization of the entropy of $\mbh\in S^D$, $S(\mbh)=-\sum_i h_i\log h_i$.
Intuitively, vectors in $S^D$ with small entropy are close to one hot encoded vectors, and thus minimizing their entropy at training time is equivalent to minimizing the KL divergence between $\mbh$ and one of the $D$ discrete degenerate distributions (Proof in \ref{sec:proofentropy}).
Our loss function is written as follows:
\begin{equation}\label{eq:loss_function}
    l(\umbx;\theta,\phi) =-\left(-\beta\sum_i h_i\log h_i+\mathbb{E}_{q_\phi(\mbz|\umbx)}\left[ \ln p_\theta(\umbx|\mbz) \right]\right),
\end{equation}
where $\beta>0$ is a tunable parameter and $\mbh=q_\phi(\mbz|\umbx)$ (Eq.~\eqref{eq:hvariable}).
Our approach has several advantages over a classical mixture of Gaussian: 
First, we do not need to specify/learn prior statistics except for the maximum number of components (=$D$). For MoG, components weight, mean and standard deviation must be specified or learnt. 
This reduces the number of trainable parameters and/or avoids certain inductive biases on the prior.
Second, this leads to a structured latent space with latent vectors directly interpretable as class probabilities, which is ideal for deep unsupervised clustering problems. 
Third, the prior is never explicitly formulated as a mixture which would require computing a bound on the KL divergence, instead we use the entropy of the vectors that is easy to compute.

The complete encoding/decoding structure of the model depicted in Fig.~\ref{fig:overallmodel} is trained on MSA data to minimize \eqref{eq:loss_function}.
At test time, the loss in \eqref{eq:loss_function} is used with $\beta=1$ as an approximation of the log-evidence in order to compute the log-likelihood ratio in \eqref{eq:logratio}.

\subsection{Reduction of the model for DMS data}\label{sec:modelreduction}
\begin{figure}[ht]
    \begin{center}
    \begin{tikzpicture}[font=\footnotesize,>=stealth,auto,node distance=1.6]
    
    \node (input) [rectangle,minimum width=1.2cm] {Label/Input};
    \node (trans_input) [rectangle, draw, right=of input] {Transformer};
    \node (mlpin) [rectangle, draw, right=of trans_input] {DwFC};
    \node (dummy) [rectangle, below=0.8cm of mlpin] {};
    \node (CE) [rectangle, draw, below=0.6cm of input] {$\text{MSE}(y,\hat{y})$};
    \node (vae) [rectangle, draw, right=of dummy] {FC};
    \node (mlpout) [rectangle, draw, below=0.8cm of dummy] {FC};
    \node (output) [rectangle, left=of trans_output,minimum width=1.2cm] {Output};
    
    \draw[->,thick] (input) -- (trans_input) node[midway, above] {$\umbx \in \mathbb{R}^{L \times d}$};
    \draw[->,thick] (trans_input) -- (mlpin) node[midway, above] {$\umbx' \in \mathbb{R}^{L \times d}$};
    \draw[->,thick] (mlpin.east) -| (vae.north)  node[midway, above] {$\umbs \in \mathbb{R}^{H\times d}$};
    \draw[->,thick] (vae.south) |- (mlpout.east) node[midway, above left] {$\mbh \in \mathbb{R}^{D}$};
    \draw[->,thick] (mlpout) -- (output.east) node[midway, above right] {};
    \draw[->,thick] (output.north) -- (CE) node[midway, right] {$\hat{y}\in\R$};
    \draw[->,thick] (input) -- (CE) node[midway, right] {$y\in\R$};
    \node[draw, dashed, fit=(trans_input) (mlpin), minimum height=1.4cm] (fit) {};
    \node[above=0cm of fit] (fit_text) {Matrix encoding};
    \end{tikzpicture}
    
    \caption{Model architecture of matENC-DMS. DwFC: Dimensionwise fully connected layer. CE: Cross-entropy loss. FC: Fully connected.}
    \label{fig:dmsdatamodel}
    \end{center}
\end{figure}
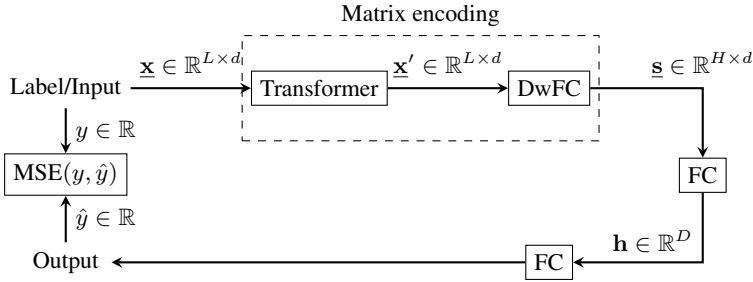

Our reduced model uses the encoding part of matVAE-MSA, and replaces the decoding part with a FC network prediction head to predict the quantitative DMS score $y\in\R$:
\begin{align*}
    \mbh &= \text{Softmax}(\text{FC}(\text{Vec}(\text{DwFC}(\text{Transformer}(\umbx)))))\in\R^D,\\
    \hat{y} &= \text{FC}(\mbh)\in\R,
\end{align*}
where $\mbh\in\R^D$ is similar to \eqref{eq:hvariable}.
The reduced model depicted in Fig.~\ref{fig:dmsdatamodel} is referred to as ``matENC-DMS".
matENC-DMS has a capacity very close to that of matVAE-MSA since the encoder is identical in the two models, and the decoder of matVAE-MSA is symmetrical to the encoder and only aims at reconstructing the input. 
We argue that this is an ideal setup to test the conservation assumption often used when designing VEPs: unfit variants were selected out by nature and thus, learning a distribution over these sequences implicitly captures the biochemical constraints that characterize fit variants.

\section{Experiments}
In this section we provide details on our experimental design choices.
Our model comparison setup with the choice of baselines and performance metrics is explained in \ref{sec:comparison}.
The model architecture and training hyper-parameters are discussed in \ref{sec:hyperparams} and \ref{sec:modeltraining}. 
The code is available: \url{https://github.com/antoinehonore/matVAE}.

\subsection{Model Comparison}\label{sec:comparison}
\paragraph{Performance metrics}
The performances of our protein specific models are measured and reported on corresponding individual DMS datasets, both in terms of Spearman’s Rank Correlation coefficient (SpearmanR) and area under receiver operating characteristic (AUROC).
The SpearmanR measures the correlation between the ranks of the predicted scores and the ranks of the target scores.
Additionally, the target score is binarized in order to compute the AUROC.
To allow for a meaningful comparison of the AUROC scores, we use the binarization threshold used in ProteinGym. 
In brief, given a DMS dataset, a threshold on the scores is selected manually between modes in case the distribution of scores is bimodal, and as the median in case the distribution of scores is unimodal. 
In the rest of the paper we will primarily compare spearmanR performances.

For matENC-DMS, we train our model on DMS data with supervised learning in a $5$-fold cross-validation framework.
Firstly, this ensures that no variant/label pair is used for both training and testing. 
Secondly, this ensures that the evaluation framework is comparable to that of matVAE-MSA, with all variants in the DMS dataset used exactly once for validation.
When a protein has multiple DMS datasets, we performed separate cross-validations on each dataset.
The performances on individual DMS datasets are then reported as the average of the $5$ models trained on independent training sets.

\paragraph{Model design and training}
We experimented with models of various latent space sizes and found that $D=10$ had best performances for MSA training. 
We kept this dimension for the rest of the experiments.
We also experimented with and without the use of the AlphaFold structures, i.e. with and without the $\text{Transformer}(.)$ layer, as well as training matEND-DMS on DMS data only, and finetuning matVAE-MSA with DMS data (matENC-FT).

\paragraph{Baselines}
All models with zero-shot prediction performances reported in the ProteinGym benchmark were considered for inclusion as a baseline.
Only those models that demonstrated the highest SpearmanR zero-shot performances on at least one pharmacogene-related protein DMS dataset were used as a baseline. 
These models fall into one of the following model families: DeepSequence \citep{riesselmanDeepGenerativeModels2018}, ESM \citep{rivesBiologicalStructureFunction2021}, Tranception \citep{notinTranceptionProteinFitness2022}, MIF \citep{laineGEMMESimpleFast2019}, GEMME \citep{laineGEMMESimpleFast2019}, VESPA \citep{marquetEmbeddingsProteinLanguage2022}, MSA Transformer \citep{raoMSATransformer2021a}.
Within each family, the top-performing models vary in configuration (e.g., different parameter counts), which we refer to as model "flavors" (See Table~\ref{tab:best_model_family}).
For example ESM2 (150M), ESM2 (15B) and ESM-1v (ensemble) are all distinct flavors within the ESM family.
We report the performances at the level of model families, by the average performances of the best performing model flavors of that family on individual DMS datasets.
The details of which model flavor performs best on which DMS dataset are shown in Table~\ref{tab:dataset_details}.
We also compare with the ``Site-Independent" model of ProteinGym, which is similar to SIFT and Polyphen-2, both still widely used in pharmacogenomics.
In addition, we add a difficult baseline referred to as ``Best Benchmark", which is the best performing model flavor across all model families for each DMS dataset (See Table~\ref{tab:best_model_family}).

To compare our models trained only on DMS data, we use the supervised learning baselines from ProteinGym with $5$ ``Random" cross-validation splits. 
For this case the models belong to the following families: ESM, TranceptEVE, Tranception, DeepSequence, MSA Transformer. 
To the best of the authors knowledge, all these baselines consist of zero-shot prediction models trained on MSA data, and used pretrained in a supervised learning framework with embeddings of the protein sequence (See \citep{notinProteinGymLargeScaleBenchmarks2023}).
ProteinNPT is a slightly different architecture which jointly trains on MSA and DMS data \citep{notin2023proteinnpt}.
The details of the best performing model flavors are in Table~\ref{tab:best_supervised_model_family}.

The prediction baseline models are summarized in Table~\ref{tab:modelcompsummary}.

\begin{table}
    \begin{center}
    \scalebox{\tablescale}{
        \begin{tabular}{|p{1.9cm}|l|p{9cm}|}
            \hline
            Source & Model Name & Short Description \\\hline
            Our & matVAE-MSA & Matrix variational auto-encoder trained on MSA data (Fig.~\ref{fig:overallmodel}) \\
            experiments & matVAE-MSA +AF & matVAE-MSA with the use of AF structures \\
             & matENC-DMS & Matrix encoder trained on DMS data (Fig.~\ref{fig:dmsdatamodel}) \\
             & matENC-DMS + AF & matENC-DMS using AF structures trained on DMS data \\
             & matENC-FT & matENC-DMS trained on MSA, and fine-tuned on DMS data \\
             & matENC-FT + AF & matENC-DMS using AF structures, trained on MSA and fine-tuned on DMS data \\\hline
            ProteinGym & ``Best Benchmark" & Best performing model flavor on a given DMS dataset \\
             & DeepSequence & VAE-based model \\
             & ESM & PLM-based model \\
             & Tranception & PLM and position-wise AA frequency-based model \\
              & TranceptEVE & PLM, EVE and position-wise AA frequency-based model \\
              & MSA Transformer & Position-wise transformer and PLM-based model \\
              & GEMME & Evolutionary tree based model\\
             & VESPA & Linear Ensemble of PLM, bio-mechanic features and positionwise frequency models.\\
             & MIF & PLM and inverse folding model\\
             & Site-Independent & Position-wise entropy-based model \\\hline
        \end{tabular}
    }
    \caption{Summary of baselines and experiments. AF stands for precomputed AlphaFold structures provided in the ProteinGym benchmark.}
    \label{tab:modelcompsummary}
    \end{center}
\end{table}

\section{Results \& Discussion}
The results for our internal experiments of our different architectures and training methods are shown in Table~\ref{tab:internal_results}.
The results are put into perspectives with average zero-shot prediction benchmark in Table~\ref{tab:our_vs_other_models} and with the supervised prediction benchmark in Table~\ref{tab:our_vs_other_models_supervised}.
The raw results of all models for zero-shot predictions are shown in Figure~\ref{fig:our_vs_other_models_spearmanr} in terms of SpearmanR and in terms of AUROC in  Figure~\ref{fig:our_vs_other_models_auroc}.
Similarly, raw results in terms of SpearmanR for supervised predictions benchmarks are shown in Figure~\ref{fig:our_vs_other_models_spearmanr_supervised}.

\subsection{Zero-shot prediction results}
When trained with MSA data only and without use of AF structures, our model (matVAE-MSA) performs better on average than "DeepSequence (single)", with an order of magnitude fewer parameters and two order of magnitude less computationally expensive inference methods\footnote{The authors of DeepSequence suggest averaging ELBO approximations over $500$ draws of latent variables (See \url{https://github.com/debbiemarkslab/DeepSequence/blob/master/examples/Mutation Effect Prediction.ipynb}), while we do not use sampling.} (Table~\ref{tab:our_vs_other_models}).
matVAE-MSA also performs better than MIF which is also a much more complex model.
In addition, matVAE-MSA performs similarly to an ensemble of DeepSequence-like architectures, which shows the improvements brought by our design to existing VAE-based VEPs.
The improvement in average performances also occurs in terms of AUROC.
We also note that these improvements are without an increase in standard deviations, which shows that our model is as reliable as others when evaluated across pharmacogene-based proteins.
Interestingly, matVAE-MSA does not benefit from the introduction of the AF structures via an input transformer model (matVAE-MSA + AF). 
The average performances decrease slightly and are comparable to MIF and "DeepSequence (single)".

\subsection{Supervised prediction results}
We evaluate our architecture in a cross-validation framework when trained using labeled DMS data, either from scratch or following pretraining with MSA data (finetuning matVAE-MSA).
First, training with DMS data increases average SpearmanR performances overall, whether or not AF structures are used in the model architecture (Table~\ref{tab:internal_results}).
However, our finetuning experiments show that our model performs poorly compared to other models, whether or not AF structures are used (Table~\ref{tab:our_vs_other_models_supervised}).
Coincidentally, both matENC-FT and matENC-DMS + AF  have identical performances, although the raw results show different performances on individual DMS datasets Figure.~\ref{fig:our_vs_other_models_spearmanr_supervised}.
When training from scratch on DMS data, the use of AF structure (matENC-DMS + AF) provides a substantial increase in performances compared to matENC-DMS, placing our model marginally above DeepSequence.
This is particularly interesting because here the DeepSequence model consists of finetuning experiments after DeepSequence is trained on MSA data.
Thus, similar performances are obtained with a simpler model and less training time.

\begin{figure}[ht]
\begin{subfigure}[t]{0.49\textwidth}
    \centering
    \includegraphics[width=\textwidth]{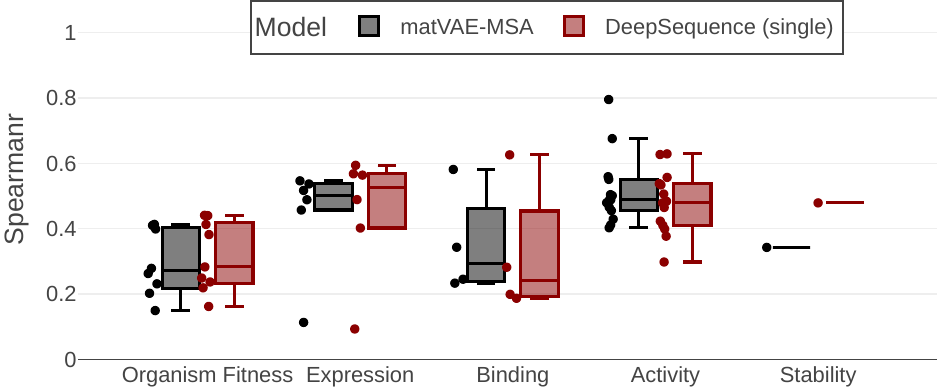}
    \caption{Raw results compared to DeepSequence}
    \label{fig:indivlatenttodms_per_coarse_selection_type}
\end{subfigure}\hfill
\begin{subfigure}[t]{0.49\textwidth}
    \centering
    \includegraphics[width=\textwidth]{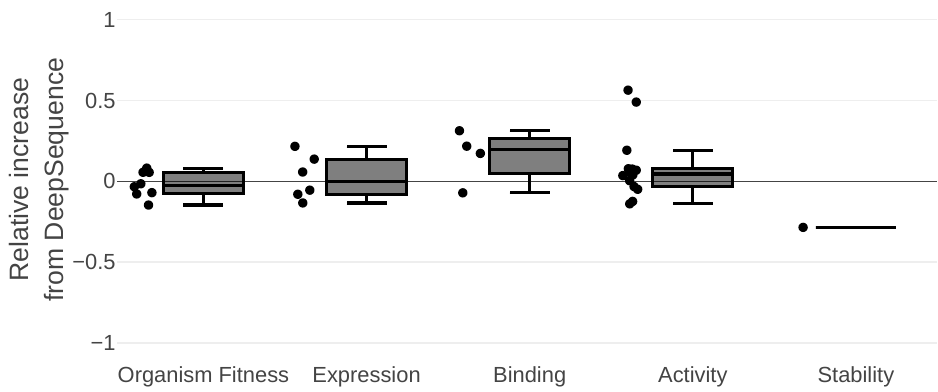}
    \caption{Relative increase in performances of matVAE-MSA compared to DeepSequence (single).}
    \label{fig:relative_increase_best_model_per_coarse_selection_type}
\end{subfigure}\\
\begin{subfigure}[t]{0.49\textwidth}
    \centering
    \includegraphics[width=\textwidth]{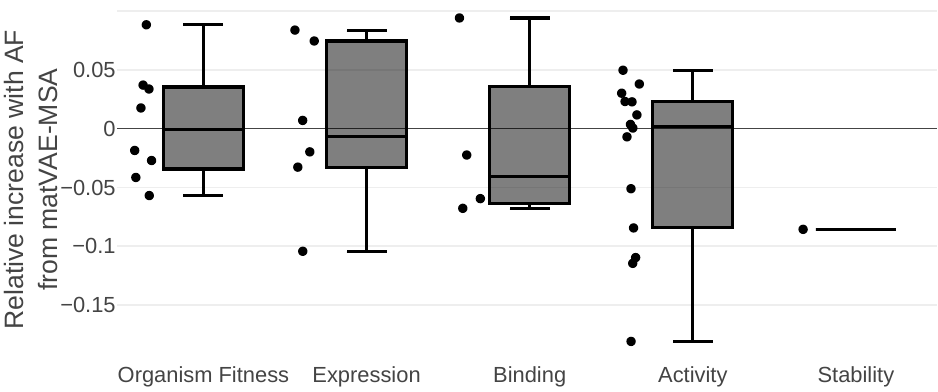}
    \caption{Relative increase with AF compared to matVAE-MSA.}
    \label{fig:relative_increase_MSAwAF_per_coarse_selection_type}
\end{subfigure}\hfill
\begin{subfigure}[t]{0.49\textwidth}
    \centering
    \includegraphics[width=\textwidth]{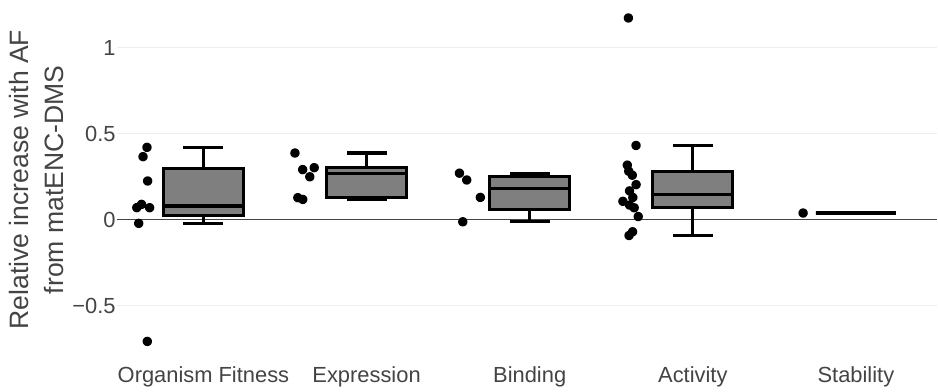}
    \caption{Relative increase with AF compared to matENC-DMS.}
    \label{fig:relative_increase_DMSwAF_per_coarse_selection_type}
\end{subfigure}
\caption{SpearmanR performances of ours models versus other zero-shot prediction models, per Selection Type sub-groups. One point denotes the score obtained on a DMS dataset and the boxplots describe the distribution of scores across DMS datasets.}
\end{figure}

When grouping DMS datasets per protein type, we find no significance in gains or loss of performances across groups (Figure.~\ref{fig:relative_increase_per_annotation}).
When grouping DMS datasets per assay selection type, we find that 
matVAE-MSA improves performances compared to DeepSequence (single) slightly more for Binding assays (Figure.~\ref{fig:relative_increase_best_model_per_coarse_selection_type}).
Interestingly, assays selecting for protein expression show a slightly more pronounced and consistent increase compared to other selection types (Figure.~\ref{fig:relative_increase_DMSwAF_per_coarse_selection_type}).

\begin{table}[ht]
\caption{Numerical results ($\mu\pm\sigma$) for our models against baselines. The models trained by us are in bold font.  AF: Use of AlphaFold structures. $^\star$ zero-shot prediction performances. $^\dagger$ supervised prediction performances. }
\begin{minipage}{0.49\textwidth}
    \begin{subtable}[t]{\textwidth}
        \begin{center}
            \scalebox{\tablescale}{
                \begin{tabular}{lll}
\toprule
 & spearmanr & auroc \\
\midrule
Best Benchmark & 0.529 $\pm$ 0.151 & 0.794 $\pm$ 0.076 \\
ESM & 0.508 $\pm$ 0.157 & 0.78 $\pm$ 0.079 \\
TranceptEVE & 0.485 $\pm$ 0.167 & 0.763 $\pm$ 0.088 \\
Tranception & 0.478 $\pm$ 0.165 & 0.761 $\pm$ 0.086 \\
GEMME & 0.461 $\pm$ 0.155 & 0.75 $\pm$ 0.085 \\
MSA Transformer & 0.452 $\pm$ 0.167 & 0.746 $\pm$ 0.089 \\
VESPA & 0.445 $\pm$ 0.164 & 0.745 $\pm$ 0.095 \\
EVE & 0.441 $\pm$ 0.165 & 0.74 $\pm$ 0.088 \\
DeepSequence (ens.) & 0.424 $\pm$ 0.152 & 0.728 $\pm$ 0.084 \\
\textbf{matVAE-MSA} & 0.423 $\pm$ 0.15 & 0.73 $\pm$ 0.08 \\
\textbf{matVAE-MSA + AF} & 0.416 $\pm$ 0.148 & 0.727 $\pm$ 0.08 \\
MIF & 0.415 $\pm$ 0.165 & 0.729 $\pm$ 0.083 \\
DeepSequence (single) & 0.412 $\pm$ 0.149 & 0.723 $\pm$ 0.083 \\
Site-Independent & 0.39 $\pm$ 0.145 & 0.713 $\pm$ 0.078 \\
\bottomrule
\end{tabular}

            }
            \caption{Zero-shot prediction performances for matVAE-MSA against baselines. DeepSequence (ens.) refers to an ensemble of DeepSequence models with various architectures.}\label{tab:our_vs_other_models}
        \end{center}
    \end{subtable}
    \end{minipage}\hfill
    \begin{minipage}{0.5\textwidth}
    \begin{subtable}[t]{\textwidth}
        \begin{center}
            \scalebox{\tablescale}{
                \begin{tabular}{ll}
    \toprule
     & spearmanr \\
    Spearmanr &  \\
    \midrule
    Best Benchmark & 0.689 $\pm$ 0.162 \\
    ProteinNPT & 0.671 $\pm$ 0.195 \\
    Tranception & 0.646 $\pm$ 0.173 \\
    ESM & 0.586 $\pm$ 0.153 \\
    MSATransformer & 0.579 $\pm$ 0.17 \\
    TranceptEVE & 0.536 $\pm$ 0.156 \\
    One-Hot Encoding & 0.528 $\pm$ 0.168 \\
    \textbf{matENC-DMS + AF} & 0.507 $\pm$ 0.183 \\
    DeepSequence & 0.501 $\pm$ 0.148 \\
    \textbf{matENC-FT} & 0.443 $\pm$ 0.187 \\
    \textbf{matENC-FT + AF} & 0.443 $\pm$ 0.187 \\
    \textbf{matENC-DMS} & 0.443 $\pm$ 0.173 \\
    \bottomrule
\end{tabular}

            }
            \caption{Supervised prediction performances for matENC-DMS against baselines.}
            \label{tab:our_vs_other_models_supervised}
        \end{center}
    \end{subtable}\\
    \begin{subtable}[t]{\textwidth}
    \centering
    \scalebox{\tablescale}{
        \begin{tabular}{|c|c|c|}
            \hline
            &  \multicolumn{2}{c|}{\textbf{SpearmanR}} \\\hline
            \textbf{Training Data} & \textbf{No AF} & \textbf{AF} \\ \hline
            MSA only$^\star$ & 0.423 $\pm$ 0.15 & 0.416 $\pm$ 0.148\\ 
            DMS only$^\dagger$ & 0.443 $\pm$ 0.173 & 0.507 $\pm$ 0.183 \\ 
            MSA $+$ DMS$^\dagger$ & 	0.443 $\pm$ 0.187 & 0.443 $\pm$ 0.187 \\ 
            \hline
        \end{tabular}
    }
    \caption{Numerical SpearmanR results on DMS datasets for different training methods. MSA $+$ DMS: Model first trained on MSA data, then finetuned with DMS data. }
    \label{tab:internal_results}
    \end{subtable}
    \end{minipage}
\end{table}

\subsection{Future work}
Our results experimenting with expressive multi-modal priors did not show improvements compared to simple Gaussian prior.
The investigations of the potential pitfalls of our current approach as well as further biology-relevant interpretations of the latent prior modes are left for future works.

Next, the joint use of both DMS and MSA data for training is an important next step towards improved model performances.
In our work, our primary objective was to evaluate the information provided by MSA and DMS data separately for variant effect prediction.
DMS and MSA data could nonetheless be used jointly for training, for instance in a fine tuning approach, where a VAE model is first trained on MSA data, and the encoding part is fine tuned on DMS data.
This might lead to improved performances since both datasets would then contribute to the model performances. 
Together with experimentation on a wider range of proteins, fine tuning on DMS data is an interesting research direction that we leave as future work.

Lastly, the design of a model able to learn from multiple proteins is also an interesting next avenue for research.
Our current architecture can be extended to work for proteins of different lengths $L_1,L_2,\dots$.
This could be done by slightly modifying the DwFC layer (Eq. \eqref{eq:DwMLP}).
An idea would be to first define $L_{M}=\max(L_1,L_2,\dots)$ and initialize $\mbU\in\R^{H\times L_M}$.
At run time, with an input protein encoding $\umbx'\in\R^{L\times d}$, the first $L$ columns from matrix $\mbU\in\R^{H\times L_M}$ can be used, which leads to: $\umbs = \mbU_{(:,1:L)} \umbx' + \mbb,$
where $\mbU_{(:,1:L)}$ denotes the sub-matrix of $\mbU$ which includes all the rows and the first $L$ columns of $\mbU$.
The complexity of the function described in \eqref{eq:DwMLP} depends on the length of the protein under consideration, while the memory complexity scales linearly with the length of the longest protein. 
This is the same as our current approach where one model is fit to individual proteins.
It differs in that the weights included up to a column $l$ would be shared between all the proteins of length at least $l$.
This makes the transformer learn to organize information, by placing what is relevant to all proteins in the first rows of the representation $\umbx'\in\R^{H\times d}$.
An issue with this approach when training on DMS data is that the meaning, support and distributions of the DMS scores vary largely across DMS datasets, thus requiring the fitness scores to be standardized (Fig.~\ref{fig:score_distribution}). 
An interesting future research direction is the quantification of similarities in selection assays of DMS datasets, so that they can exploited in regression models.

\section{Conclusion}
We proposed a transformer-based matrix variational auto-encoder together with a structured prior distribution. 
We evaluated the performances of the model on a variant effect prediction using DMS and MSA datasets from the ProteinGym benchmark
When training on MSA data (matVAE-MSA) and performing zero-shot predictions on the DMS dataset, our proposed approach outperforms the state-of-art DeepSequence model.
Our model achieves this with an order of magnitude fewer parameters, and less computations at inference time. 
Our architecture allowed to compare performances with models of similar capacity but trained on DMS datasets (matENC-DMS) instead of MSA.
We then trained our architecture solely on DMS.
DMS datasets were often much smaller than MSA data for the same protein.
matENC-DMS still outperformed matVAE-MSA on average for supervised prediction tasks.
We saw further increase in performances when using alphafold generated structures in our transformer model, which lead to similar performances compared to DeepSequence pretrained on MSA and finetuned on DMS data. 
This additionally shows the efficacy of our method.
DMS assays may thus replace MSAs entirely, without major losses in performances.
This in turn motivates the development of DMS datasets and the study of their relationships, in order to further improve variant effect prediction.

\bibliography{main}
\bibliographystyle{iclr2025_conference}

\newpage
\renewcommand\thefigure{\thesection.\arabic{figure}}
\renewcommand\thetable{\thesection.\arabic{table}}    

\appendix
\section{Appendix}
\subsection{Entropy minimization of discrete latent vectors}\label{sec:proofentropy}
\def\p{\mathbb{P}}
Let $d\in\N$, and $\forall i\in[d]\quad\alpha_i>0$.
Suppose $\alpha=(\alpha_1,\dots,\alpha_d)$ is the parameter vector of a Dirichlet distribution with $d$ categories.
Let $X\sim \text{Dir}(\alpha)$.
Let $Z$ a one-of-d binary vector such that $\forall i \in[d] \quad\p(Z_i=1,Z_{j\neq i}=0)=X_i$.
We have the conditional entropy: 
\begin{equation}
H(Z|X)=\mathbb{E}[-\ln \p(Z|X)]=-\sum_{i=1}^dX_i\ln X_i.
\end{equation}
Let $\p_U$ denote a discrete uniform distribution with a support of size $d$.
We also have that 
\begin{equation}\label{eq:entropyKLequivalence}
    H(Z|X)=\ln d -\mathbb{E}_X[D_{KL}(\p(Z|X)||\p_U(Z))].
\end{equation}
\begin{proof}
By definition (Cover \& Thomas) we have: $H(Z|X) = H(Z) - I(Z;X)$.
To prove Eq.~\eqref{eq:entropyKLequivalence}, the idea is to introduce the uniform discrete distribution in the terms on the RHS:\begin{align}
    H(Z)&=-\sum_i \p(Z_i)\ln \p(Z_i)\\
    &=-\sum_i \p(Z_i)\ln \frac{\p(Z_i)\p_U(Z_i)}{\p_U(Z_i)}\\
    &=-\sum_i \p(Z_i)\ln\p_U(Z_i) -\sum_i \p(Z_i)\ln \frac{\p(Z_i)}{\p_U(Z_i)}\\
    &=\ln d - D_{KL}(\p(Z)||\p_U(Z))
\end{align}

Similarly, we introduce the uniform discrete distribution in the mutual information:
\begin{align}
    I(Z;X) &= D_{KL}\left(\p(Z,X)||\p(Z)\p(X)\right)\\
    &=\sum_{z,x}\p(z,x)\ln \frac{\p(z,x)}{\p(z)\p(x)}\\
    &=\sum_{z,x}\p(z,x)\ln \frac{\p(z,x)\p_U(z)}{\p(z)\p(x)\p_U(z)}\\
    &=\sum_{z,x}\p(z,x)\ln \left[\frac{\p(z,x)}{\p(x)\p_U(z)}+\frac{\p_U(z)}{\p(z)}\right]\\
    &=D_{KL}\left(\p(Z,X)||\p_U(Z)\p(X)\right)-\sum_z \ln\frac{\p(z)}{\p_U(z)}\sum_x\p(z,x)\\
    &=\mathbb{E}_X\left[D_{KL}\left(\p(Z|X)||\p_U(Z)\right)\right] - D_{KL}\left(\p(Z)||\p_U(Z)\right)
\end{align}
Subtracting the terms concludes the proof:\begin{equation}
    H(Z|X) = \ln d - \mathbb{E}_X\left[D_{KL}\left(\p(Z|X)||\p_U(Z)\right)\right]
\end{equation}
\end{proof}
Therefore decreasing $H(Z|X)$ increases $\mathbb{E}_X[D_{KL}(\p(Z|X)||\p_U(Z))]$, i.e. in the limit, \begin{equation}
    H(Z|X)\rightarrow 0\quad\implies\quad\p(Z|X)\rightarrow Q(Z)
\end{equation}
where $Q(Z)$ is the one-of-$d$ discrete distribution with entropy $H(Q)=0$.

\subsection{Datasets}\label{sec:dataset}
We train separate models on $26$ pharmacogene-related proteins for which the DMS and MSA datasets are readily available from the publicly available ProteinGym repository \citep{notinProteinGymLargeScaleBenchmarks2023}.
The pharmacogene-related proteins were divided into four functional categories: Drug Targets ($n=21$) and Absorption-Distribution-Metabolism-Excretion (ADME) related proteins ($n=5$).
The ADME category is further divided into Cytochrome (``CYP"), ``transporter" and ``other" ADME proteins. In total we compare performances on $33$ DMS datasets, some proteins having several DMS datasets obtained under different selection assays (Table~\ref{tab:dataset_details}).

\subsubsection{Preprocessing of MSA sequences}
We followed the preprocessing steps proposed in DeepSequence for the MSA data  \citep{riesselmanDeepGenerativeModels2018}.
Sequences are removed from MSAs if they include more than $50\%$ gaps.
Columns are removed from a MSA if they contain more than $30\%$ gaps across the MSA.
For consistency, the columns that are removed from the MSA are also removed from the DMS datasets of that protein.
When training on MSAs, each sequence is sampled with a probability proportional to the reciprocal of the number of sequences within a given Hamming distance from that sequence \citep{riesselmanDeepGenerativeModels2018}.
A summary description of the datasets is provided in Table~\ref{tab:datasetsummary}.

\begin{table}[ht!]
    \begin{center}
    \caption{Description of DMS and MSA datasets per protein category. The most extreme values are in bold. L: Preprocessed sequence length; MSA Num Seq (resp. DMS Num Seq): number of sequences in the MSA (resp. DMS) datasets.
     ADME trans.: ADME Transporter.}\label{tab:datasetsummary}
    \scalebox{\tablescale}{
        \begin{tabular}{|cc|ccc|}
\hline
Category &  & L & MSA Num Seq & DMS Num Seq \\\hline
{ADME CYP (n=2)} & $\mu\pm\sigma$ & 490 $\pm$ 0 & 260849 $\pm$ 0 & 6256 $\pm$ 161 \\
 & min/max & 490 / 490 & 260849 / 260849 & 6142 / 6370 \\\hline
{ADME other (n=2)} & $\mu\pm\sigma$ & 204 $\pm$ 57 & 86361 $\pm$ 94716 & 3246 $\pm$ 569 \\
 & min/max & 164 / 245 & 19387 / 153335 & 2844 / 3648 \\\hline
{ADME tran. (n=3)} & $\mu\pm\sigma$ & 579 $\pm$ 44 & 144978 $\pm$ 90553 & 10491 $\pm$ 951 \\
 & min/max & 553 / 630 & 40416 / 197259 & 9803 / 11576 \\\hline
{Drug target (n=26)} & $\mu\pm\sigma$ & 498 $\pm$ 427 & 62331 $\pm$ 125352 & 3374 $\pm$ 3134 \\
 & min/max & \textbf{31} / \textbf{1863} & \textbf{911} / \textbf{611225} & \textbf{63} / \textbf{12464} \\
\hline
\end{tabular}
}
    \end{center}
\end{table}

\subsection{Model architectures hyper-parameters}\label{sec:hyperparams}
The encoding and decoding transformers of matVAE-MSA are designed with $3$ transformer layers described in \eqref{eq:transformerlayer}. 
The embedding dimensions of all the layers are identical and equal to $d$.
The use of $3$ layers allows to use information from neighbors up to order $3$ according to the graph defined by the attention mask.
The attention mask is a thresholded distance matrix derived from the WT protein structures predicted by Alphafold2 \citep{jumperHighlyAccurateProtein2021a}.
This means that queries are allowed to attend to keys in the attention dot product if the predicted distance in $3$d between the corresponding AAs is $\leq c$.
We chose $c=7$\AA~which had the best performances for 4 out of 5 graph neural network-based models predicting variant effects in \citep[Table S3]{gelmanNeuralNetworksLearn2021}.
The structures are readily available in the ProteinGym repository as PDB files \citep{notinProteinGymLargeScaleBenchmarks2023}.
For DwFC, we chose $H=\min(H_{min},L)$ with $H_{min}=200$.
This means that the dimension is not reduced for proteins with small enough sequence length $L\leq H_{min}$.
Following the discussion in Section \ref{sec:qr}, we experimented with $H_{min}\approx d$ on some proteins, but could not get good performances.
For FCB, we used a $2$-layer ReLU network with $1000$ and $300$ neurons, and output latent space dimension $D=10$, this is smaller than the design of DeepSequence which uses $D=50$ \citep{riesselmanDeepGenerativeModels2018}.
Symmetrical design choices were used for the decoding part of matVAE-MSA.
For training, we use $\beta=0.01$ and at test time (i.e. when computing the ELBO) we use $\beta=1$.
We found that larger $\beta$ caused the latent vectors to collapse to identical vectors, and smaller $\beta$ caused the training to be unconstrained, i.e. leading to uniform vectors.
For matENC-DMS, the $D$-dimensional latent vector in the FC bottleneck is passed into a prediction head with a $1$-layer fully connected ReLU network, with $10$ neurons, and an output dimension of $1$. 
The decoding parts of matVAE-MSA are not used.

\subsection{Model training}\label{sec:modeltraining}
For matVAE-MSA, the models are trained on protein specific MSAs to minimize the negative ELBO in \eqref{eq:elbo}. 
The expected reconstruction error is approximated with a 1-sample Monte Carlo method. 
The reconstruction error is the cross-entropy between the true $\umbx\in\R^{L\times d}$ and the reconstructed $\hat{\umbx}$. 
For matENC-DMS, no variational formulation is used. 
The loss function is the mean squared error (MSE) between the true label $y\in\R$ and the reconstructed label $\hat{y}$. 

For all our included proteins, the loss function for matVAE-MSA is optimized using the ADAM optimizer, with a fixed learning rate $\lambda=8e-5$, a batch size of $256$ and $300,000$ training steps.
For matENC-DMS, the loss function is optimized with a fixed learning rate $\lambda=1e-4$, a batch size of $512$ and $100,000$ training steps.
The learning rates were chosen similar to the optimal one reported for a graph neural network model in \citep[Table~S3]{gelmanNeuralNetworksLearn2021}. 
The batch sizes were chosen to obtain the most efficient use of our hardware.
In matVAE-MSA, the memory footprint is mainly due to the attention matrices in the encoder and decoder transformer. 
We double the batch size for matENC-DMS compared to matVAE-MSA since matENC-DMS only has an encoder transformer.


\subsection{Raw SpearmanR and AUROC performances for zero-shot learning tasks}
\begin{figure}[ht]
    \begin{center}
        \includegraphics[width=\textwidth]{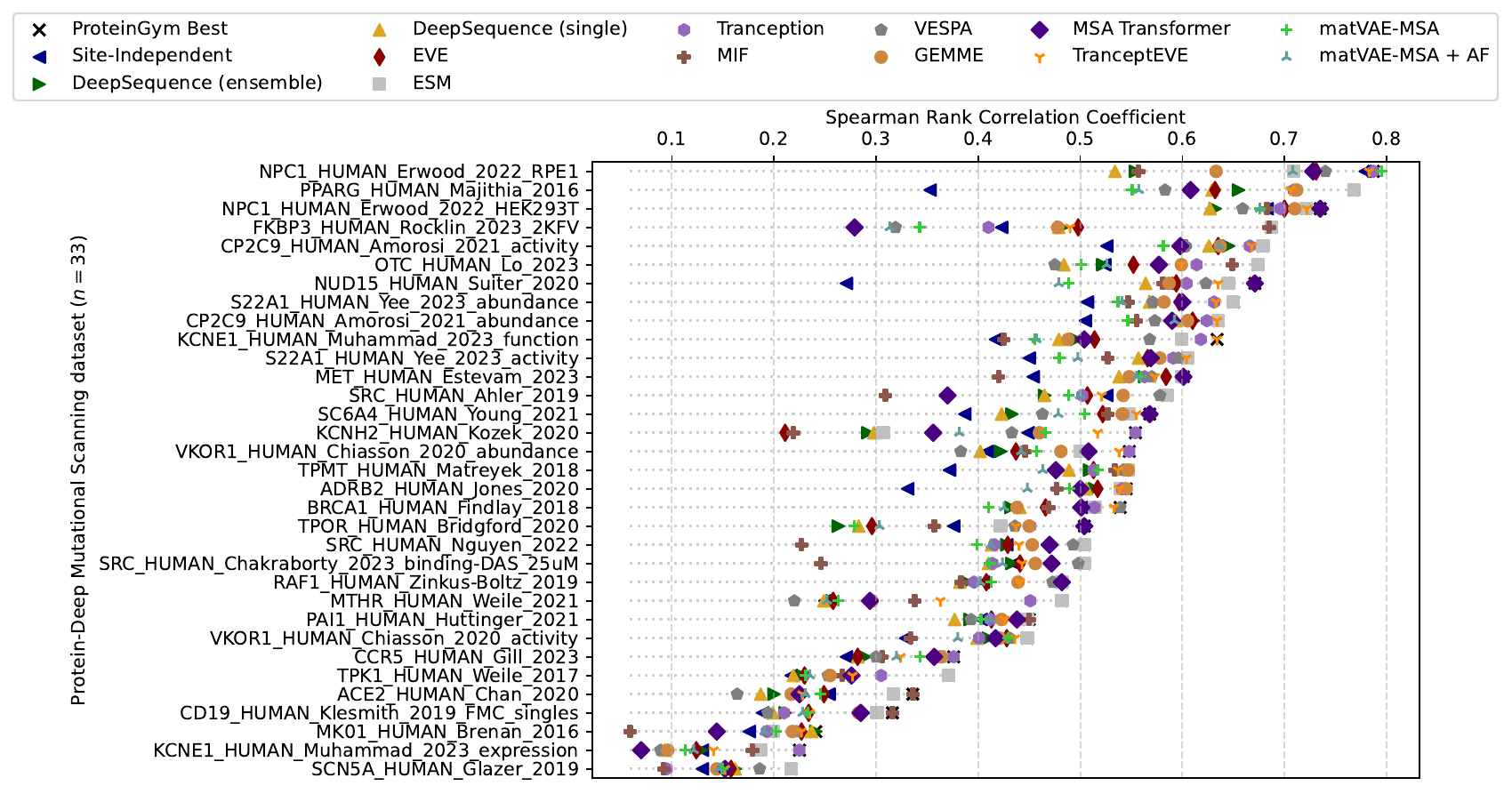}
        \caption{Raw Spearman Rank correlation coefficient performances on all DMS datasets for zero-shot prediction task. We display the performances of our models against the best performing models in ProteinGym for pharmacogene-related proteins. The datasets are sorted in decreasing ``best benchmark" performances.}\label{fig:our_vs_other_models_spearmanr}
    \end{center}
\end{figure}

\begin{figure}[ht]
    \begin{center}
    \includegraphics[width=\textwidth]{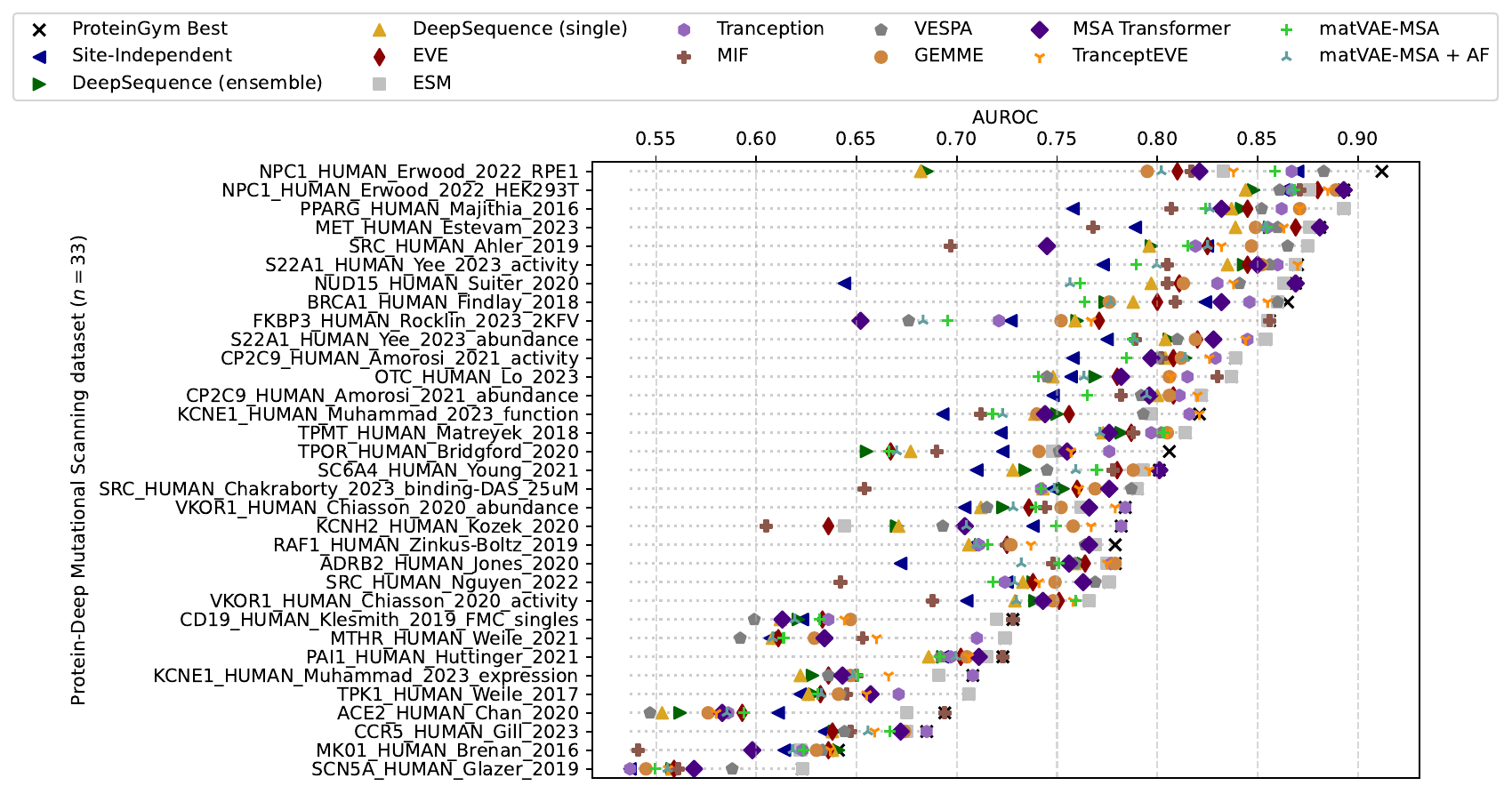}
    \end{center}
    \caption{AUROC performances on all DMS datasets for zero-shot prediction task. We display the performances of our models against the best performing models in ProteinGym for pharmacogene-related proteins. The datasets are sorted in decreasing ``best benchmark" performances.}
    \label{fig:our_vs_other_models_auroc}
\end{figure}

\begin{figure}[ht]
    \begin{center}
        \includegraphics[width=\textwidth]{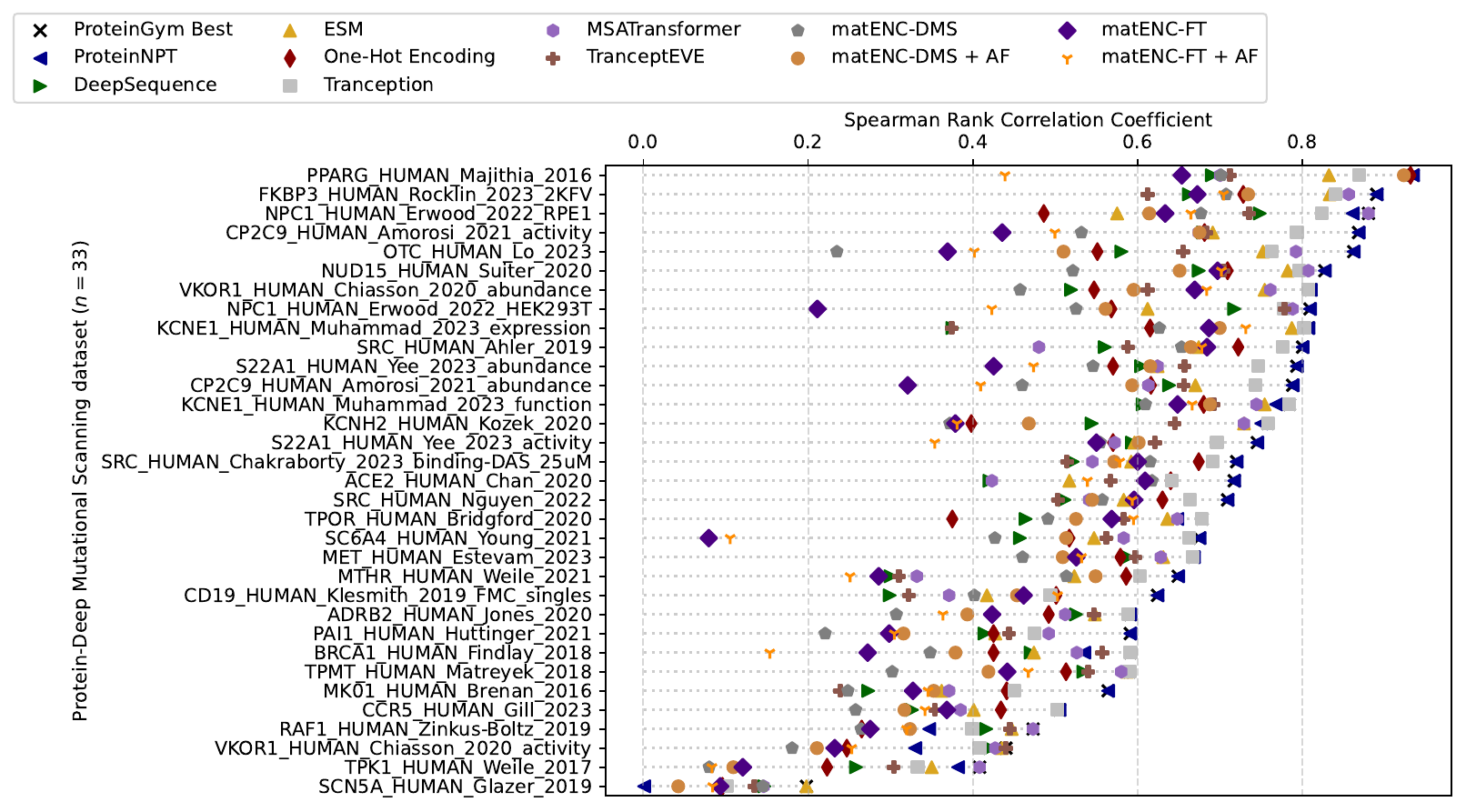}
    \end{center}
    \caption{SpearmanR performances on all DMS datasets for a supervised learning task. We display the performances of our models against the best performing models in ProteinGym for pharmacogene-related proteins. The datasets are sorted in decreasing ``best benchmark" performances.}
    \label{fig:our_vs_other_models_spearmanr_supervised}
\end{figure}
\newpage
\subsection{Performances according to protein category}

\begin{figure}[ht!]
\begin{subfigure}{0.49\textwidth}
    \centering
    \includegraphics[width=\textwidth]{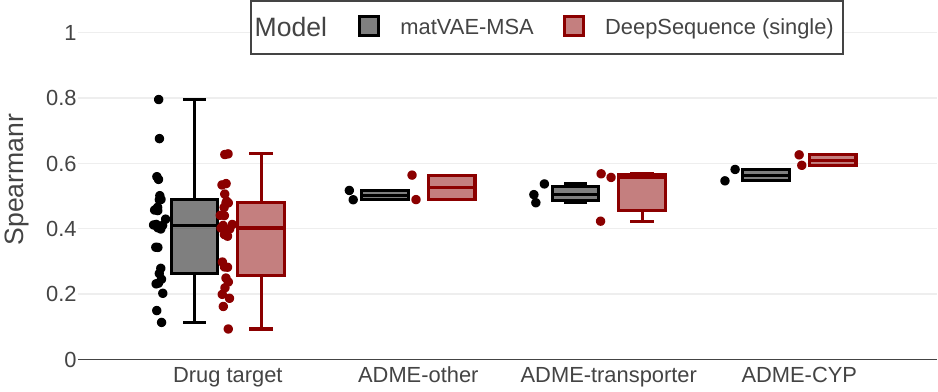}
    \caption{Raw results for matVAE-MSA (black) and ``Best Benchmark" (red).}
    \label{fig:indivlatenttodms_per_annotation}
\end{subfigure}\hfill
\begin{subfigure}{0.49\textwidth}
    \centering
    \includegraphics[width=\textwidth]{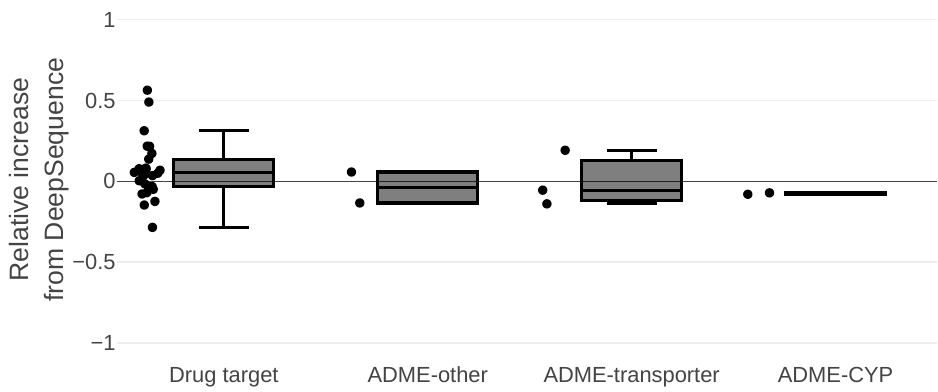}
    \caption{Relative increase of matENC-DMS compared to ``Best Benchmark".}
    \label{fig:relative_increase_best_model_per_annotation}
\end{subfigure}\\
\begin{subfigure}{0.49\textwidth}
    \centering
    \includegraphics[width=\textwidth]{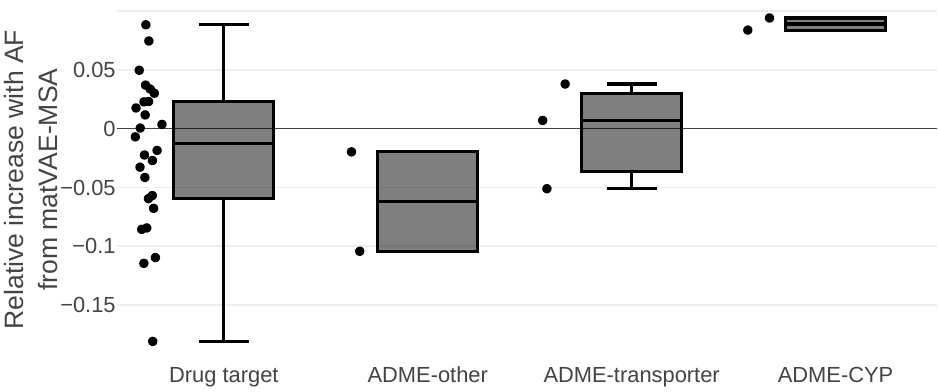}
    \caption{Relative increase of matVAE-MSA+AF compared to matVAE-MSA.}
    \label{fig:relative_increase_MSA_per_annotation}
\end{subfigure}\hfill
\begin{subfigure}{0.49\textwidth}
    \centering
    \includegraphics[width=\textwidth]{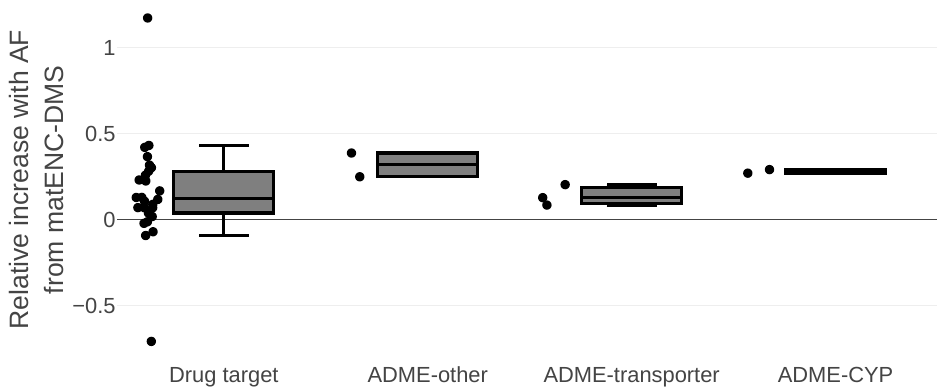}
    \caption{Relative increase of matENC-DMS+AF compared to matENC-DMS.}
    \label{fig:relative_increase_enc_per_annotation}
\end{subfigure}
\caption{SpearmanR performances of matENC-DMS versus other zero-shot prediction models, per protein category. One point denotes the score obtained on a DMS dataset and the boxplots describe the distribution of scores across DMS datasets.}
\label{fig:relative_increase_per_annotation}
\end{figure}

\newpage
\subsection{Studying possible confounder for relative increase of finetuning from MSA}
\begin{figure}[ht]
    \begin{subfigure}{0.49\textwidth}
        \includegraphics[width=\textwidth]{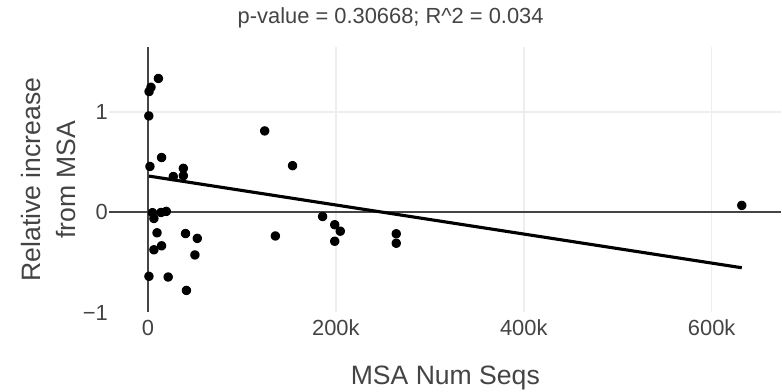}
        \caption{}\label{fig:relative_increase_MSA_vs_MSA_num_seqs}
    \end{subfigure}\hfill
    \begin{subfigure}{0.49\textwidth}
        \includegraphics[width=\textwidth]{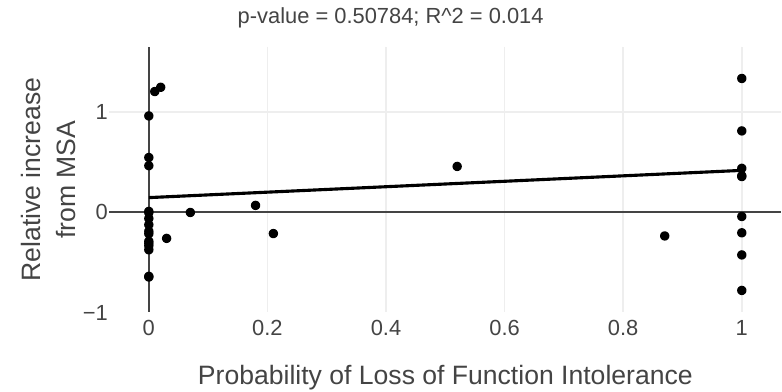}
        \caption{}\label{fig:relative_increase_MSA_vs_pLI}
    \end{subfigure}\\
        \begin{subfigure}{0.49\textwidth}
        \includegraphics[width=\textwidth]{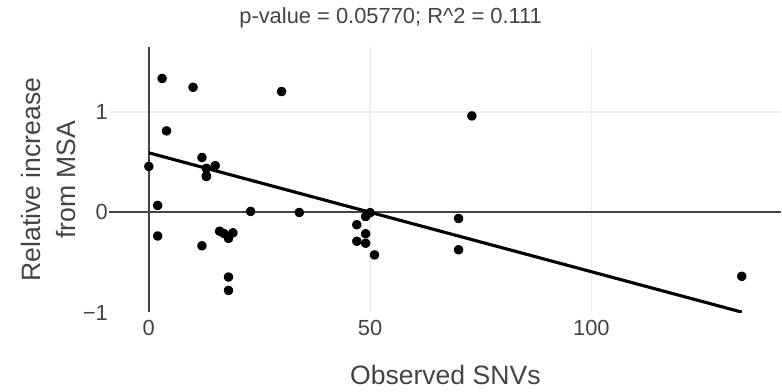}
        \caption{}\label{fig:relative_increase_MSA_vs_Observed_SNVs}
    \end{subfigure}\hfill
    \begin{subfigure}{0.49\textwidth}
        \includegraphics[width=\textwidth]{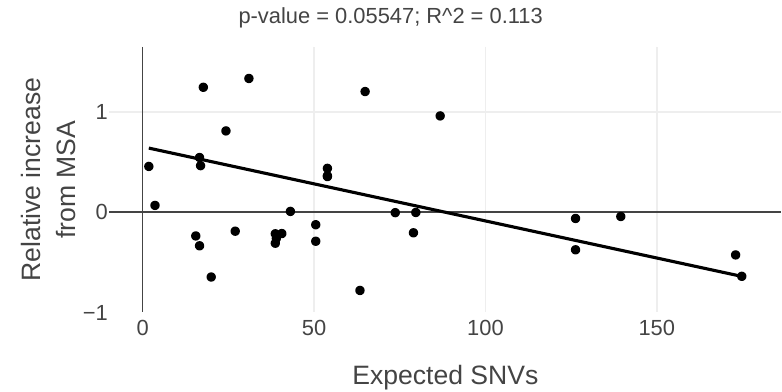}
        \caption{}\label{fig:relative_increase_MSA_vs_Expected_SNVs}
    \end{subfigure}\\
    \begin{subfigure}{0.49\textwidth}
        \includegraphics[width=\textwidth]{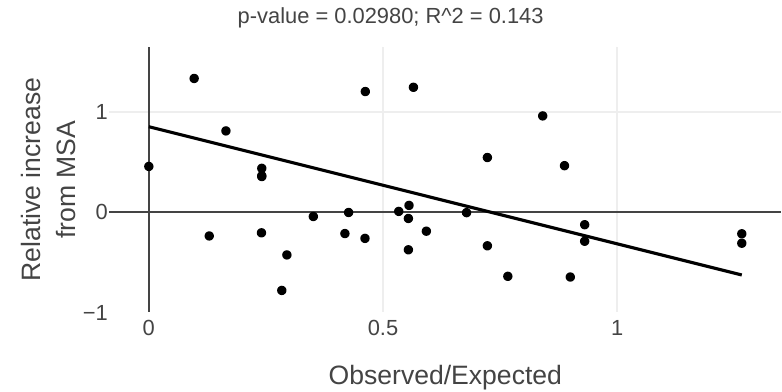}
        \caption{}\label{fig:relative_increase_MSA_vs_oe}
    \end{subfigure}\hfill
    \begin{subfigure}{0.49\textwidth}
        \includegraphics[width=\textwidth]{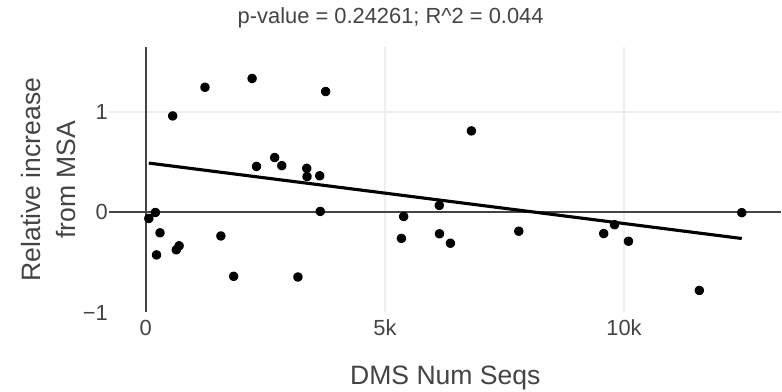}
        \caption{}\label{fig:relative_increase_MSA_vs_DMS_total_number_mutants}
    \end{subfigure}
    \caption{Univariate correlation analysis for potential confounders of the relative increase of matENC-DMS + AF from matVAE-MSA + AF. None of the considered confounders are significant to explain the relative increase from MSA. A multivariate correlation analysis was performed and did not show any significance (not shown). \textbf{MSA Num Seqs}: Number of variants in MSA; \textbf{Probability of Loss of Function Intolerance}: Genes with a pLI close to 1 are  often associated with haploinsufficiency and dominant genetic diseases;
    \textbf{Expected (resp. Observed) SNVs}: Expected (resp. Observed) Single-nucleotide variant in each gene;
    \textbf{Observed/Expected (o/e)}: Constrained genes have fewer observed variants than expected (low o/e) and are under a higher degree of selection than less constrained genes.
    \textbf{DMS Num Seqs}: Number of variants in DMS data; 
    }\label{fig:relative_increase_MSA_vs_cofounders}
\end{figure}

\newpage
\subsection{Additional dataset information}

\begin{figure}[ht!]
    \centering
    \includegraphics[width=0.8\textwidth]{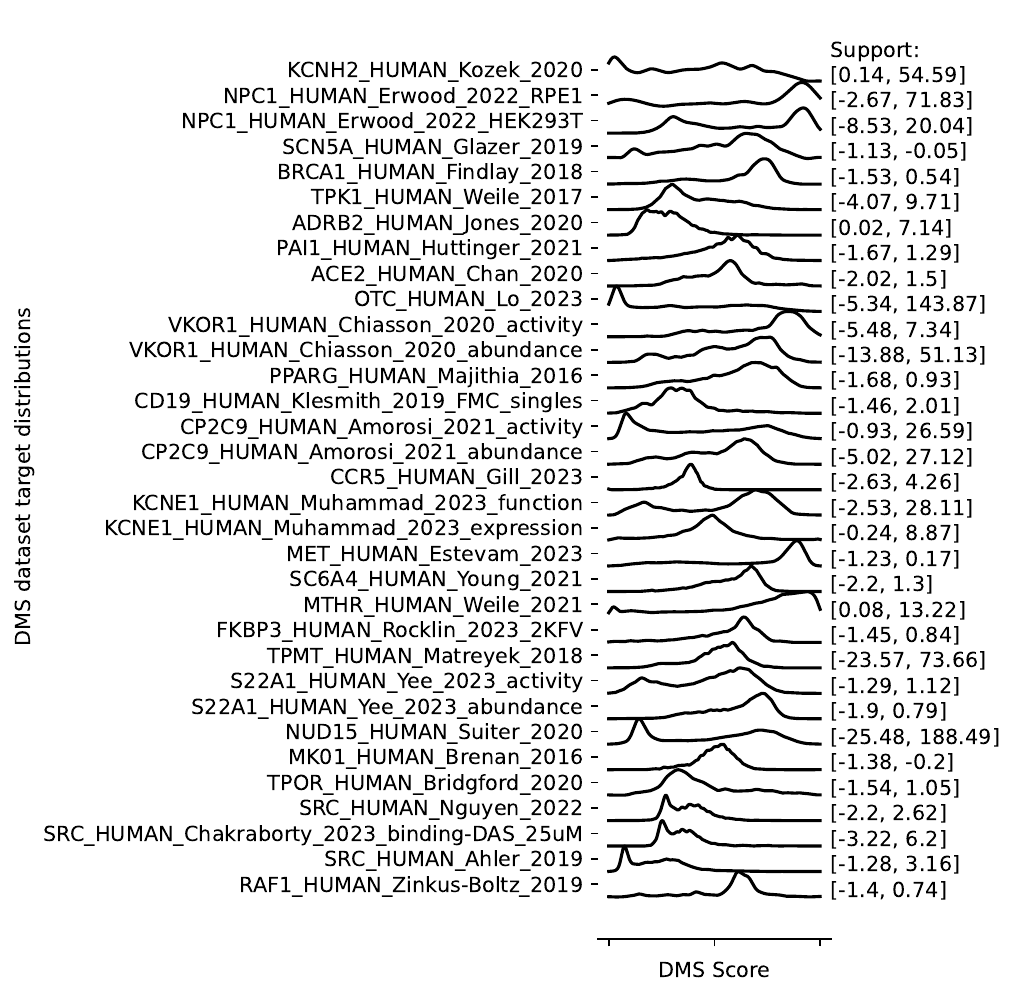}
    \caption{Empirical kernel density estimates of the density function of DMS datasets target score. The support and range of each density function are standardized to [0,1] for visualization. The true support of the density functions is annotated on the right hand side. The densities are displayed with an increment of $1$ along the y-axis.}
    \label{fig:score_distribution}
\end{figure}

\begin{table}[ht!]
    \begin{center}
    \scalebox{\tablescale}{
        \begin{tabular}{|p{1.6cm}|p{1.6cm}|p{0.7cm}|p{1.2cm}|p{1.2cm}|p{2cm}|p{4.5cm}|}
\hline
\textbf{Category} & \textbf{UniProt ID} & \textbf{L} & \textbf{MSA Num Seq} & \textbf{DMS Num Seq} &  \textbf{Selection Type} & \textbf{Best Benchmark - Flavor}  \\
\hline
ADME CYP & CP2C9 & 490 & 260849 & 6142 & Binding & ESM2 (150M) \\
 &  &  &  & 6370 & Expression & ESM2 (650M) \\\hline
ADME other & NUD15 & 164 & 153335 & 2844 & Expression & MSA Transformer (ensemble) \\
 & TPMT & 245 & 19387 & 3648 & Expression & ESM-1v (ensemble) \\\hline
ADME transporter & S22A1 & 553 & 197259 & 9803 & Expression & ESM-1v (ensemble) \\
 &  &  &  & 10094 & Activity & ESM-1v (ensemble) \\
 & SC6A4 & 630 & 40416 & 11576 & Activity & MSA Transformer (ensemble) \\\hline
Drug target & ACE2 & 805 & 10865 & 2223 & Binding & MIF \\
 & ADRB2 & 413 & 201108 & 7800 & Activity & GEMME \\
 & BRCA1 & 1863 & 974 & 1837 & Organismal Fitness & VESPA \\
 & CCR5 & 352 & 611225 & 6137 & Binding & Tranception L \\
 & CD19 & 556 & 1171 & 3761 & Binding & MIF \\
 & FKBP3 & 69 & 3211 & 1237 & Stability & ESM-IF1 \\
 & KCNE1 & 129 & 2104 & 2315 & Activity & TranceptEVE L \\
 &  &  &  & 2339 & Expression & Tranception M no retrieval \\
 & KCNH2 & 31 & 13900 & 200 & Activity & Tranception M \\
 & MET & 287 & 184827 & 5393 & Activity & MSA Transformer (ensemble) \\
 & MK01 & 360 & 123422 & 6809 & Organismal Fitness & DeepSequence (ensemble) \\
 & MTHR & 656 & 4724 & 12464 & Organismal Fitness & ESM2 (150M) \\
 & NPC1 & 1278 & 6234 & 63 & Activity & Tranception S \\
 &  &  &  & 637 & Activity & MSA Transformer (ensemble) \\
 & OTC & 354 & 134484 & 1570 & Activity & ESM-IF1 \\
 & PAI1 & 402 & 51792 & 5345 & Activity & MIF-ST \\
 & PPARG & 505 & 39639 & 9576 & Activity & ESM2 (15B) \\
 & RAF1 & 648 & 9609 & 297 & Organismal Fitness & MSA Transformer (single) \\
 & SCN5A & 32 & 49959 & 224 & Organismal Fitness & ESM-1v (single) \\
 & SRC & 536 & 37311 & 3366 & Organismal Fitness & ESM-1v (ensemble) \\
 &  &  &  & 3372 & Activity & ESM-1v (ensemble) \\
 &  &  &  & 3637 & Activity & ESM-1v (ensemble) \\
 & TPK1 & 243 & 21338 & 3181 & Organismal Fitness & ESM2 (15B) \\
 & TPOR & 635 & 911 & 562 & Organismal Fitness & MSA Transformer (single) \\
 & VKOR1 & 163 & 14425 & 697 & Activity & ESM-1v (ensemble) \\
 &  &  &  & 2695 & Expression & Tranception L \\
\hline
\end{tabular}

    }
    \end{center}
    \caption{Deep Mutation Scanning datasets details retrieved from ProteinGym. \textbf{Uniprot ID}: Universal protein identifier; \textbf{L}: Preprocessed sequence length; \textbf{MSA Num Seq} (resp. \textbf{DMS Num Seq}): number of sequences in the MSA (resp. DMS) datasets. \textbf{Selection Type}: DMS assay selection type. \textbf{Best Benchmark - Flavor}: best performing model flavor for zero-shot prediction tasks.}
    \label{tab:dataset_details}
\end{table}
\begin{table}[ht]
    \begin{center}
    \scalebox{\tablescale}{
        \begin{tabular}{l|l}
\hline
 \textbf{Best Benchmark - Family} & \textbf{Best Benchmark - Flavor} \\
\hline
ESM (n=15) & ESM2 (15B) x2 \\
 & ESM-1v (ensemble) x7 \\
 & ESM2 (150M) x2 \\
 & ESM2 (650M) x1 \\
 & ESM-IF1 x2 \\
 & ESM-1v (single) x1 \\\hline
MSA Transformer (n=6) & MSA Transformer (single) x2 \\
 & MSA Transformer (ensemble) x4 \\\hline
Tranception (n=5) & Tranception L x2 \\
 & Tranception S x1 \\
 & Tranception M x1 \\
 & Tranception M no retrieval x1 \\\hline
MIF (n=3) & MIF-ST x1 \\
 & MIF x2 \\\hline
DeepSequence (n=1) & DeepSequence (ensemble) x1 \\\hline
TranceptEVE (n=1) & TranceptEVE L x1 \\\hline
GEMME (n=1) & GEMME x1 \\\hline
VESPA (n=1) & VESPA x1 \\
\hline
\end{tabular}

    }
    \end{center}
    \caption{Summary of best model flavors and families for zero-shot prediction tasks.}
    \label{tab:best_model_family}
\end{table}
\begin{table}[ht!]
    \begin{center}
    \scalebox{\tablescale}{
        \begin{tabular}{ll}
\toprule
 & Best Benchmark Name \\
Model Family &  \\
\midrule
ProteinNPT (n=23) & ProteinNPT x23 \\
Tranception (n=5) & Tranception Embeddings x5 \\
MSA Transformer (n=2) & MSA Transformer Embeddings x2 \\
ESM (n=1) & ESM-1v + One-Hot Encodings x1 \\
MSA Transformer + One-Hot Encodings (n=1) & MSA Transformer + One-Hot Encodings x1 \\
TranceptEVE (n=1) & TranceptEVE + One-Hot Encodings x1 \\
\bottomrule
\end{tabular}

    }
    \end{center}
    \caption{Summary of best model flavors and families for supervised prediction tasks.}
    \label{tab:best_supervised_model_family}
\end{table}

\end{document}